\documentclass[sigconf]{acmart}

\usepackage{booktabs} 
\usepackage{multirow}
\usepackage{flushend}
\setcopyright{acmcopyright}

\copyrightyear{2019} 
\acmYear{2019} 
\acmConference[CIKM '19]{The 28th ACM International Conference on Information and Knowledge Management}{November 3--7, 2019}{Beijing, China}
\acmBooktitle{The 28th ACM International Conference on Information and Knowledge Management (CIKM '19), November 3--7, 2019, Beijing, China}
\acmPrice{15.00}
\acmDOI{10.1145/3357384.3358100}
\acmISBN{978-1-4503-6976-3/19/11}

\begin{document}
\fancyhead{}

\title{Large Margin Prototypical Network for Few-shot Relation Classification with Fine-grained Features}

\author{Miao Fan, Yeqi Bai, Mingming Sun, Ping Li}
\affiliation{Cognitive Computing Lab\\ Baidu Research\\
No.10 Xibeiwang East Road, Beijing, 10085, China\\
10900 NE 8th St, Bellevue, WA 98004, USA}


\email{{fanmiao, v_baiyeqi, sunmingming01, liping11}@baidu.com}

\begin{abstract}
Relation classification (RC) plays a pivotal role in both natural language understanding and knowledge graph completion. It is generally formulated as a task to recognize the relationship between two entities of interest appearing in a free-text sentence. Conventional approaches on RC, regardless of feature engineering or deep learning based, can obtain promising performance on categorizing common types of relation leaving a large proportion of unrecognizable long-tail relations due to insufficient labeled instances for training. In this paper, we consider few-shot learning is of great practical significance to RC and thus improve a modern framework of metric learning for few-shot RC. Specifically, we adopt the large-margin ProtoNet with fine-grained features, expecting they can generalize well on long-tail relations. Extensive experiments were conducted by \textit{FewRel}, a large-scale supervised few-shot RC dataset, to evaluate our framework: \textit{LM-ProtoNet (FGF)}. The results demonstrate that it can achieve substantial improvements over many baseline approaches. 
\end{abstract}

%
%



\maketitle

\section{Introduction}
Relation classification (RC)~\citep{10.1007/978-981-10-7359-5_6} is a pivotal task for both natural language understanding (NLU)~\citep{hirschberg2015advances} and knowledge graph completion (KGC)~\citep{nickel2016review}. Given a free-text sentence, we can first adopt an off-the-shelf software (e.g., Stanford CoreNLP~\citep{P14-5010}) to discover entities of interest, and RC is a successive module which takes in charge of recognizing the relationship between each pair of the named entities. In this way, we can extract many triplets, denoted as ({\it head-entity, relation, tail-entity}), from free-text sentences. These triplets can facilitate machines understanding the inner structure of natural language and even extending knowledge bases. 

A great number of approaches have been proposed in the past decades for the task of sentence-level RC, including feature engineering methods~\citep{zelenko2003kernel,culotta2004dependency} and neural learning models~\citep{Sun:2018:LUE:3159652.3159712,sun-etal-2018-logician}. Regardless of those ways of producing various evidence for RC, conventional approaches mainly rely on a large number of labeled instances for each type of relation in the training phase, and the promising performance of RC is merely obtained on common relations. Many manually annotated corpora of RC such as MUC-7 and ACE can only cover $1\%$-$2\%$ of the relations recorded by some large-scale knowledge bases, such as Freebase~\citep{bollacker2008freebase}, DBpedia~\citep{lehmann2015dbpedia} and Wikidata~\citep{42240} where thousands of relations are included. 

\begin{table*}
    \begin{center}
    \caption{An example of a $5$-way-$1$-shot scenario of RC. For each sentence, the head entity (in blue) and the tail entity (in red) are indicated in advance. Given a query instance, the few-shot RC model is responsible for selecting the correct relation (from R1 to R5) expressed by the instance, according to the support set.}
    \vspace{-3mm}
        \resizebox{0.9\textwidth}{!}
        {
        \begin{tabular}{l|c}
            \toprule
            \multicolumn{2}{c}{\textbf{Support Set}}
            \\
            \hline
            R1: \textit{head of government}
            & The 1926 {\color{blue} United States} elections were held in President {\color{red} Calvin Coolidge}'s second term. \\
            R2: \textit{author of} 
            & "{\color{blue} Heaven Help Us All}" is a 1970 soul single composed by {\color{red} Ron Miller}. \\
            R3: \textit{country of} 
            &  Daisy Geyser is a geyser of {\color{red} Yellowstone National Park} in the {\color{blue} United States.} \\
            R4: \textit{position held} & It is led by {\color{blue} Viorica Dncil}, who assumed office as {\color{red} Prime Minister of Romania} on 29 January 2018. \\
            R5: \textit{architect} 
            & {\color{blue}Capital Gate} was designed by architectural firm {\color{red} RMJM} and was completed in 2011. \\
            \hline
            \hline
            \multicolumn{2}{c}{\textbf{Query Instance}}
            \\
            \hline
            R1, R2, R3, R4, or R5 
            & {\color{blue} Belgium}'s highest point is the {\color{red} Signal de Botrange} at 694 meters above the sea level. \\
            \bottomrule
        \end{tabular}
        }

\vspace{-4mm}
            \label{tab:example}
    \end{center}
\end{table*}

To alleviate the issue of insufficient labeled instances for long-tail relations, \citet{P09-1113} proposed the \textit{distant supervision} paradigm to automatically label free-text sentences with the relations from an existing knowledge base by some heuristic alignment rules\footnote{An intuitive but straightforward rule is that if two entities participate in a relation, any sentence that contain those two entities might express that relation.} to build training data. It, however, naturally suffers from the problem of incorrect annotations.

The emerging study on few-shot learning~\cite{Fei-Fei:2006:OLO:1115692.1115783,chen19closerfewshot} inspires us that we can train an RC model with the instances labeled by common relations and transfer the knowledge to infer more sentences that may express long-tail relations. Few-shot RC models are expected to obtain considerable accuracy supported by very few instances annotated by long-tail relations without learning from scratch. For example, Table~\ref{tab:example} shows an example of a $5$-way-$1$-shot scenario of RC, where we can leverage a few-shot RC model $\mathcal{M}$ to tell whether the query instance/sentence mentions any type of the $5$ relations given by the support set ($1$ shot/instance for each relation). What makes few-shot RC unique is that the model $\mathcal{M}$ is trained by the instances labeled by other relations where the $5$ relations are not included.

To attract more successive studies on few-shot RC, \citet{han2018fewrel} constructed a large-scale supervised few-shot relation classification dataset (\textit{FewRel})
for the purpose of evaluating the performance of various meta-learning approaches, including Meta Network~\citep{munkhdalai2017meta}, GNN~\citep{garcia2018fewshot}, SNAIL~\citep{mishra2018a} and ProtoNet~\citep{NIPS2017_6996}, on RC. They reported that the CNN-based~\citep{C14-1220} ProtoNet outperforms the other baseline methods. However, we consider the framework can be further improved either by task-specific features or by advanced learning targets.

In this paper, we contribute two updates to the original CNN-based~\citep{C14-1220} ProtoNet from the perspectives of fine-grained feature generation and large-margin learning, respectively, aiming at increasing the generalization ability of few-shot RC models on recognizing long-tail relations. Further experiments were also conducted on \textit{FewRel}, and the results demonstrate that our framework, i.e. \textit{LM-ProtoNet (FGF)}, attains a leading performance over many baselines with a substantial improvement by \textbf{6.84\%} accuracy.
\begin{figure*}[!htp]
\centering
\includegraphics[width=0.9\textwidth]{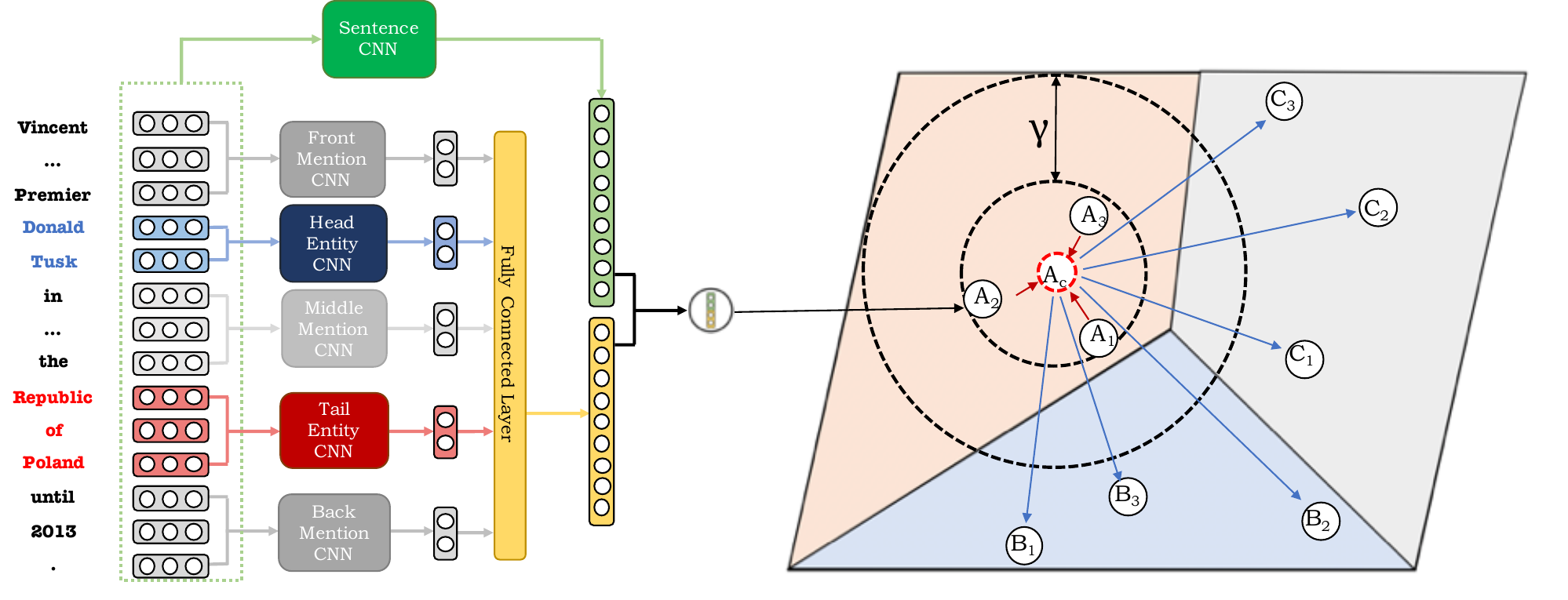}
        \vspace{-4mm}
\caption{The framework of {\it LM-ProtoNet (FGF)} which is composed of two modules: fine-grained features for instance embedding (on the left) and triplet loss for ProtoNet (on the right). In this case, {\it LM-ProtoNet} is addressing a $3$-way (classes: $A$, $B$, and $C$) -$3$-shot RC, and $A_c$ is the center of class $A$.} 
        \vspace{-3mm}
\label{fig:framework}
\end{figure*} 

\section{Proposed Framework}
Figure~\ref{fig:framework} illustrates our {\it LM-ProtoNet (FGF)} framework for few-shot RC. It is composed of two updates on improving the original ProtoNet~\citep{NIPS2017_6996} from the perspectives of the feature generation and the learning target.

\subsection{Fine-grained Features}
A simple way of encoding an instance/sentence is to adopt a CNN~\citep{C14-1220} to generate a fix-length embedding to represent the semantics of the whole sentence. However, we consider that the recognized entities of interest can provide additional benefits on producing fine-grained features. 
As shown by the left panel in Figure 1, we employ multiple CNNs to generate a phrase-level embedding besides the sentence-level embedding as the feature to represent the input sentence $x$. Therefore, the encoding $f(x)$ of input sentence $x$ can be formulated as:
\begin{equation}
f(x) = f_{sentence}(x) \oplus f_{phrase}(x),
\end{equation}
where $\oplus$ is the operator to concatenate the sentence-level and the phrase-level embeddings to generate fine-grained features. 

Specifically speaking, $f_{sentence}(x)$ denotes the way of encoding the sentence $x$ via a conventional CNN~\citep{C14-1220} parameterized by $\Theta$ for RC, which can be expressed as:
\begin{equation}
    f_{sentence}(x) = \text{CNN}(x;\Theta).
\label{eq_cnn}
\end{equation}
The phrase-level network $f_{phrase}$ leverages finer granularity of the sentence $x$ which is segmented into \textit{five} parts/phrases: the relation mention in the front $r_f$, the head entity $e_h$, the relation mention in the middle $r_m$, the tail entity $e_t$, and the relation mention in the back $r_b$. Each of the phrases can be encoded by the CNN in Equation~\ref{eq_cnn} and here we use the bold fonts $\textbf{r}_f$, $\textbf{e}_h$, $\textbf{r}_m$, $\textbf{e}_t$, and $\textbf{r}_b$, to denote their embeddings. The phrase-level network takes the CNN embeddings of these phrases as inputs and feeds them into a fully connected layer activated by the ReLU~\citep{DBLP:journals/corr/XuWCL15} function parameterized by $\Phi$ to explore the non-linear relationship between these features:
\begin{equation}
f_{phrase}(x) = \text{ReLU}(\textbf{r}_f \oplus \textbf{e}_h \oplus \textbf{r}_m \oplus \textbf{e}_t \oplus \textbf{r}_b;\Phi).
\end{equation}

\begin{table*}[!htp]
    \begin{center}
    \caption{Accuracies (\%) of all models for evaluating on the \textit{FewRel} test set under four different settings.}
    \vspace{-3mm}
        \begin{tabular}{l|cccc}
            \toprule
            \textbf{Few-shot RC Model} & \textbf{5 Way 1 Shot} & \textbf{5 Way 5 Shot} & \textbf{10 Way 1 Shot} & \textbf{10 Way 5 Shot} \\
            \hline
            \hline
            
            Finetune (CNN) & $44.21\pm0.44$ & $68.66\pm0.41$ & $27.30\pm0.28$ & $55.04\pm0.31$ \\
            Finetune (PCNN) & $45.64\pm0.62$ & $57.86\pm0.61$ & $29.65\pm0.40$ & $37.43\pm0.42$ \\
            KNN (CNN) & $54.67\pm0.44$ & $68.77\pm0.41$ & $41.24\pm0.31$ & $55.87\pm0.31$ \\
            KNN (PCNN) & $60.28\pm0.43$ & $72.41\pm0.39$ & $46.15\pm0.31$ & $59.11\pm0.30$ \\
            \hline
            
            Meta Network (CNN) & $64.46\pm0.54$ & $80.57\pm0.48$ & $53.96\pm0.56$ & 
            $69.23\pm0.52$\\
            GNN (CNN)
            & $66.23\pm0.75$ & $81.28\pm0.62$ & 
            $46.27\pm0.80$ & $64.02\pm0.77$ \\
            SNAIL (CNN) & $67.29\pm0.26$ & $79.40\pm0.22$ & $53.28\pm0.27$ & $68.33\pm0.26$ \\
            
            ProtoNet (CNN) & $69.20\pm0.20$ & $84.79\pm0.16$ & $56.44\pm0.22$ & $75.55\pm0.19$ \\
            
            LM-ProtoNet (FGF) & $\textbf{76.60}\pm\textbf{0.24}$ & $\textbf{89.31}\pm\textbf{0.13}$ & $\textbf{65.31}\pm\textbf{0.31}$ & $\textbf{82.10}\pm\textbf{0.21}$ \\
            \hline
           Human Performance & $92.22\pm5.53$ & - & $85.88\pm7.40$ & - \\
            \bottomrule
            
        \end{tabular}
                    \vspace{-4mm}
            \label{tab:result}
    \end{center}
\end{table*}

\begin{table*}[!htp]
    \begin{center}
    
        \caption{Accuracies (\%) of all models for ablation study on the \textit{FewRel} validation set under four different settings.}  
        \vspace{-3mm}
        \begin{tabular}{l|cccc}
            \toprule
            \textbf{Few-shot RC Model} & \textbf{5 Way 1 Shot} & \textbf{5 Way 5 Shot} & \textbf{10 Way 1 Shot} & \textbf{10 Way 5 Shot} \\
            \hline
            \hline
            ProtoNet (CNN)
            & $68.40\pm0.34$ & $84.28\pm0.29$ & 
            $54.47\pm0.12$ & $71.26\pm0.45$ \\
            ProtoNet (FGF) 
            & $71.62\pm0.15$ & $85.13\pm0.21$ & $60.26\pm0.34$ & $74.58\pm0.19$ \\
            LM-ProtoNet (CNN) 
            & $71.02\pm0.23$ & $84.29\pm0.11$ & $60.74\pm0.18$ & $74.27\pm0.21$ \\
            LM-ProtoNet (FGF) 
            & $73.24\pm0.17$ & $86.38\pm0.23$ & $62.05\pm0.22$ & $76.97\pm0.16$ \\
            \bottomrule
            
        \end{tabular}

        \vspace{-4mm}
            \label{tab:dis}

    \end{center}
\end{table*}

\subsection{Triplet Loss for Large Margin ProtoNet}
The performance of prototypical network for RC highly depends on the spacial distribution of sentence embeddings. Therefore, we added an auxiliary loss function to force both the sentence network and the phrase network to enlarge the distances between classes and shorten the distances within the same class. We adopt the triplet loss as the additional learning target:
\begin{equation}
\small
\mathcal{L}_{triplet} = \sum_{i=1}^{N} \max{(0, \gamma + ||f(a_i) - f(p_i)||^2 - ||f(a_i) - f(n_i)||^2)}, 
\end{equation}
where $N$ is the total number of training episodes and $ (a_i, p_i, n_i) $ is a triplet consists of the anchor, the positive, and the negative instance.
In ProtoNet, the anchor is a virtual instance, and its embedding is the average of the instances in a support set sampled from the training set.  

The final loss $\mathcal{L}$ is a trade-off between the softmax cross-entropy loss $\mathcal{L}_{softmax}$ and the triplet loss $\mathcal{L}_{triplet}$ adapted by a hyper-parameter $\lambda$:
\begin{equation}
    \mathcal{L} = \mathcal{L}_{softmax} + \lambda * \mathcal{L}_{triplet}.
\end{equation}

\section{Experiments}
\subsection{Benchmark Dataset}
            
            
            
To the best of our knowledge, {\it FewRel}~\citep{han2018fewrel} is the first large-scale annotated corpus for the task of sentence-level few-shot relation classification. This benchmark dataset was built by the subsequent two steps. The distant supervision was first adopted to align the sentences in Wikipedia and the triplets in Wikidata to generate a candidate set of $122,000$ instances automatically labeled by $122$ relations. To filter out the incorrect labels, the human annotation then involved in and $100$ relations ($700$ instances for each relation) were reserved. To create a scenario of few-shot RC, {\it FewRel} has separated subsets for training, validation, and testing, covered by $64$, $16$, and $20$ relations, respectively. And the relations that occur in each subset must be mutually exclusive. 
\subsection{Experimental Settings}
The comparison methods are carefully selected for the sake of covering almost all the representative approaches on RC.
To be specific, they are the non-parametric KNN, the deep learning approaches (CNN~\citep{C14-1220} and PCNN~\cite{zeng2015distant}), and several up-to-date meta-learning models including Meta Network~\citep{munkhdalai2017meta}, GNN~\citep{garcia2018fewshot}, SNAIL~\citep{mishra2018a} and ProtoNet~\citep{NIPS2017_6996}. 
We fine-tuned the hyper-parameters of our framework {\it LM-ProtoNet (FGF)} via observing its accuracy on the validation set and evaluated the performance of the best model on the test set.

\subsection{Result Analysis}
Table 3 reports the accuracies of all the mentioned approaches on the {\it FewRel} test set. The results indicate that performance of CNN~\citep{C14-1220} and PCNN~\citep{zeng2015distant} dramatically decrease as we only have $1$ or $5$ instances for each new type of relation to fine-tune these deep neural networks, which is far from adequate for generalization. The non-parametric KNN can generalize better than the deep models in the few-shot scenario with a great leap forward on accuracy. The meta-learning approaches generally obtain better results and ProtoNet~\citep{NIPS2017_6996} achieves the highest accuracy among them. {\it LM-ProtoNet (FGF)} updates both the way of feature generation and the learning target of the original ProtoNet and obtains a leading performance with a significant improvement by \textbf{6.84\%} accuracy.
\begin{figure}[!htp]
\centering
\includegraphics[width=0.48\textwidth]{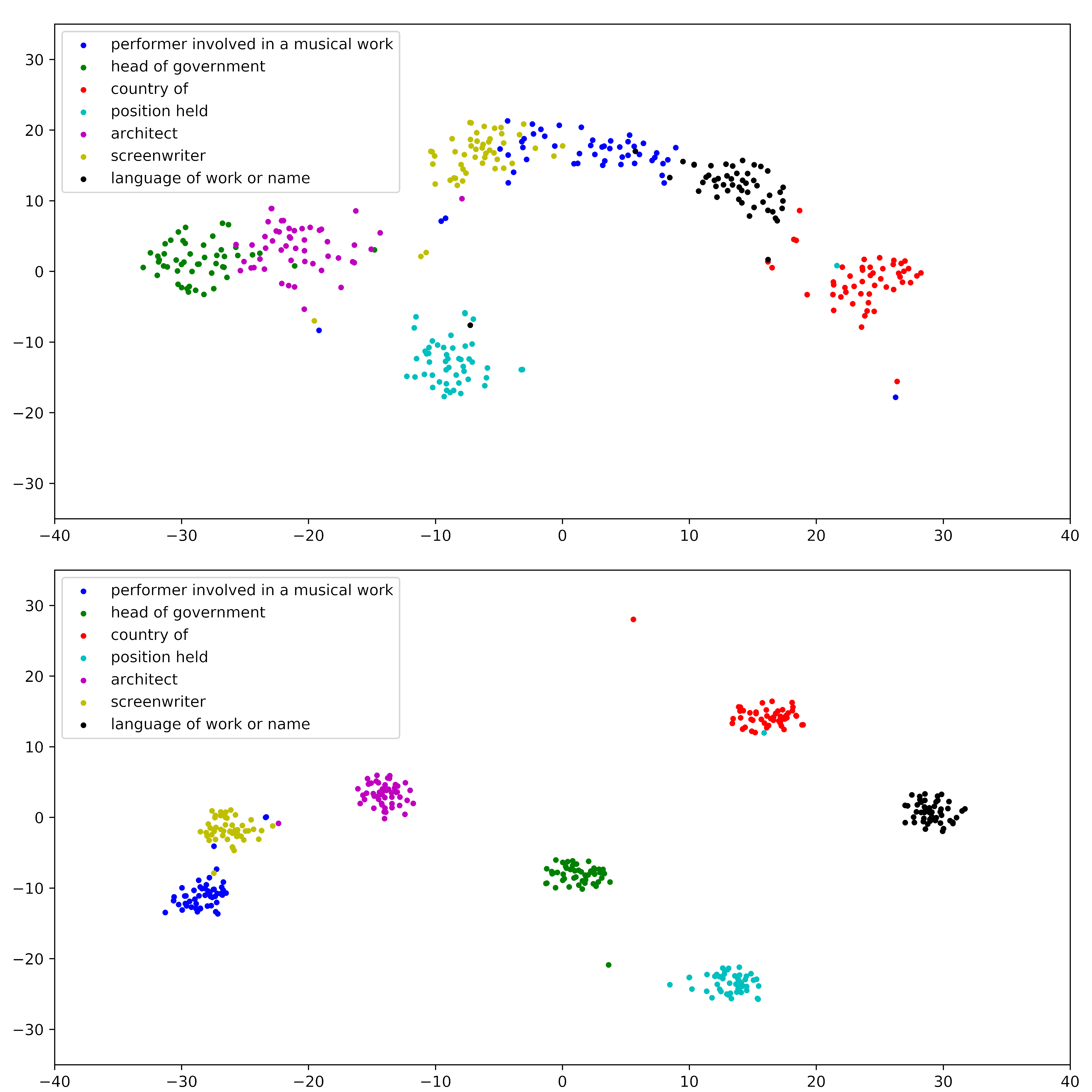}
                    \vspace{-6mm}
\caption{a $7$-way-$40$-shot scenario of RC where the embeddings of the instances in the support set are acquired by ProtoNet on the top and {\it LM-ProtoNet (FGF)} at the bottom. The embeddings are mapped into the same 2D metric space by the technique of t-SNE.}
                    \vspace{-3mm}
\label{fig:dis_2}
\end{figure} 

\section{Ablation Studies}
In this section, we will explore the effectiveness of each update we have proposed. Given the restrictions of access to the test set of {\it FewRel}, we split the original validation set into two parts, leaving $8$ relations as the test set for ablation studies. 

\subsection{Fine-grained Features v.s. CNN-based Embedding}
The intuition for using fine-grained features (\textit{FGF}) instead of CNN-based embedding is that we can measure the distance between two instances by additional discriminative evidence such as the entities of interest and the context around them. To verify our assumption, we apply \textit{FGF} into ProtoNet and LM-ProtoNet, respectively. Table~\ref{tab:dis} shows that using \textit{FGF} can consistently boost the accuracy by \textbf{2.69\%}, regardless of the metric learning methods we adopt.

\subsection{Triplet Loss v.s. Softmax Cross-entropy}
The essence of employing the triplet loss as a new learning target for ProtoNet is to maintain a larger margin for the long-tail relations which do not occur in the training set. We believe it can increase the generalization ability of our few-shot RC model. With the help of t-SNE \cite{maaten2008visualizing}, we map the embeddings of the instances in a $7$-way-$40$-shot scenario into a 2D metric space.  As illustrated by Figure~\ref{fig:dis_2}, \textit{LM-ProtoNet} acquires more discriminative representations for instances by leaving ample room in the metric space. Table~\ref{tab:dis} also demonstrates that \textit{LM-ProtoNet} can consistently improve the accuracy by \textbf{2.37\%}, despite of the way of generating features. 

\section{Conclusion}
Few-shot RC is emerging research for information extraction given its promising capability of recognizing new types of relationships with very few labeled instances. 
This paper improves a state-of-the-art framework of metric learning for the task of few-shot relation classification (RC), via contributing two updates from the perspectives of fine-grained feature generation and large-margin learning, respectively. The purpose of our framework is to increase the generalization ability of few-shot RC models on recognizing long-tail relations.
Extensive experiments were conducted on \textit{FewRel}, a newly published dataset to evaluate methods on large-scale supervised few-shot RC. The results show that \textit{LM-ProtoNet (FGF)} obtains significant improvements in accuracy.

\bibliographystyle{ACM-Reference-Format}
\bibliography{sample-bibliography}

\end{document}